%% file: safecbarxivv2.tex
\documentclass{article}

\usepackage{PRIMEarxiv}
\usepackage{natbib}

\input{neurips2023/macros}
\input{neurips2023/title}
\input{neurips2023/authors}

\begin{document}
\maketitle

\input{neurips2023/abstract}

\input{neurips2023/introduction}
\input{neurips2023/problemsetting}
\input{neurips2023/algorithms}
\input{neurips2023/experiments}
\input{neurips2023/relatedwork}

\input{neurips2023/discussion}

\bibliographystyle{plainnat}  
\bibliography{safecbarxivv2.bbl}

\newpage
\input{neurips2023/appendix}

\end{document}

%% file: neurips2023/macros.tex
\usepackage[utf8]{inputenc}

\usepackage{microtype}
\usepackage{graphicx}
\usepackage{booktabs} 

\usepackage{hyperref}

\usepackage{amsmath}
\usepackage{amssymb}
\usepackage{mathtools}
\usepackage{amsthm}

\usepackage[capitalize,noabbrev]{cleveref}

\theoremstyle{plain}
\newtheorem{theorem}{Theorem}[section]

\newtheorem{lemma}[theorem]{Lemma}
\newtheorem{corollary}[theorem]{Corollary}
\theoremstyle{definition}

\theoremstyle{remark}
\newtheorem{remark}{Remark}[section]

\usepackage[textsize=tiny]{todonotes}

\PassOptionsToPackage{colorlinks = true, linkcolor = blue, urlcolor  = blue, citecolor = blue, anchorcolor = blue}{hyperref}

\usepackage[utf8]{inputenc} 
\usepackage[T1]{fontenc}    
\usepackage{hyperref}       
\usepackage{url}            
\usepackage{booktabs}       
\usepackage{amsfonts}       
\usepackage{nicefrac}       
\usepackage{microtype}      

\usepackage{xcolor}         
\usepackage{soul}           
\usepackage{amsmath}
\usepackage{graphicx}
\usepackage{xcolor}
\usepackage{algorithm}
\usepackage[noend]{algorithmic}
\usepackage{transparent} 
\usepackage{xspace}
\usepackage{amsthm}
\usepackage{thmtools}
\usepackage{thm-restate}
\usepackage{comment}
\usepackage{wrapfig}
\usepackage{multirow}
\usepackage{subcaption} 
\usepackage{footnote}
\usepackage{tabularx}
\usepackage{afterpage}

\newcommand{\AlgOpt}{\ensuremath{\mathrm{\mathbf{Alg}}_{\mathsf{Opt}}}\xspace}
\newcommand{\AlgReg}{\ensuremath{\mathrm{\mathbf{Alg}}_{\mathsf{Reg}}}\xspace}
\newcommand{\RegCB}{\ensuremath{\mathrm{\mathbf{Reg}}_{\mathsf{CB}}}\xspace}
\newcommand{\RegEVaRq}{\ensuremath{\mathrm{\mathbf{Reg}}_{\EVaR_q}}\xspace}
\newcommand{\VaR}{\ensuremath{\operatorname{\mathsf{VaR}}}\xspace}
\newcommand{\CVaR}{\ensuremath{\operatorname{\mathsf{CVaR}}}\xspace}
\newcommand{\predict}{\operatorname{\mathsf{predict}}}
\newcommand{\update}{\operatorname{\mathsf{update}}}
\newcommand{\EVaR}{\ensuremath{\operatorname{\mathsf{EVaR}}}\xspace}

\newcommand{\fastcb}{\textsf{FastCB}\xspace}
\newcommand{\squarecb}{\textsf{SquareCB}\xspace}
\newcommand{\smoothcb}{\textsf{SmoothCB}\xspace}
\newcommand{\corral}{\textsf{Corral}\xspace}
\newcommand{\scope}{\textsf{Scope}\xspace}

\newcommand{\abelong}{\ensuremath{\operatorname{\mathsf{AL}}}\xspace}
\newcommand{\contabelong}{\ensuremath{\operatorname{\mathsf{Cont-AL}}}\xspace}

\DeclareMathOperator\erf{erf}

\newcommand{\R}{\ensuremath{\mathbb{R}}\xspace}
\newcommand{\Rinf}{\ensuremath{\R \cup \{\infty\}}\xspace}
\newcommand{\cA}{\ensuremath{\mathcal{A}}\xspace}
\newcommand{\cX}{\ensuremath{\mathcal{X}}\xspace}

\newcommand{\repourl}{\texttt{\url{https://github.com/zwd-ms/risk\_averse\_cb}}}

\usepackage{tablefootnote}

%% file: neurips2023/title.tex
\title{Conditionally Risk-Averse Contextual Bandits}

%% file: neurips2023/authors.tex
\author{
  M\'{o}nika Farsang \\
  Vienna University of Technology \\
  \texttt{monika.farsang@tuwien.ac.at} \\
  \And
  Paul Mineiro \\
  Microsoft Research \\
  \texttt{pmineiro@microsoft.com} \\
  \And
  Wangda Zhang \\
  Microsoft Research \\
  \texttt{wangdazhang@microsoft.com}
}

%% file: neurips2023/abstract.tex
\begin{abstract}
Contextual bandits with average-case statistical guarantees are inadequate in risk-averse situations because they might trade off degraded worst-case behaviour for better average performance.  Designing
a risk-averse contextual bandit is challenging because exploration is necessary but risk-aversion is sensitive to the entire distribution of rewards; nonetheless we exhibit the first risk-averse contextual bandit algorithm with an online regret guarantee. We conduct experiments from diverse scenarios where worst-case outcomes should be avoided, from dynamic
pricing, inventory management, and self-tuning software; including a production exascale data processing system.

\end{abstract}

%% file: neurips2023/introduction.tex
\section{Introduction}
Contextual bandits~\citep{auer2002nonstochastic,langford2007epoch}
are a mature technology with numerous applications:
however, adoption has been most aggressive in recommendation
scenarios~\citep{DBLP:journals/corr/abs-1904-10040}, where the
worst-case outcome is user annoyance.  At the other extreme are medical
and defense scenarios where worst-case outcomes are literally fatal.
In between are scenarios of interest where bad outcomes are tolerable
but should be avoided, e.g., logistics; finance; and self-tuning
software, where the term \emph{tail catastrophe} highlights the
inadequacy of average case performance guarantees in real-world
applications~\citep{marcus2021bao}. These scenarios demand risk-aversion,
i.e., decisions should sacrifice average performance in order to avoid
worst-case outcomes, and incorporating risk-aversion into contextual
bandits would facilitate adoption.
More generally, risk aversion is essential for making informed decisions that align with the risk preferences of the decision maker by balancing the potential benefits and risks of a particular action. 

This paper solves risk-averse decision making for contextual bandits
via reduction to regression, resulting in the \emph{first risk-averse
contextual bandit algorithm with an online regret guarantee}.  The regret
guarantee applies over adversarially chosen context sequences and
includes the exploration choices made by the algorithm.  The approach
utilizes arbitrary (online learnable) function classes and extends to
infinite action spaces; introduces no computational overhead relative
to the risk-neutral setting; introduces statistical overhead directly
related to the desired level of risk-aversion, with no overhead in
the risk-neutral limit; and composes with other innovations within the
Decision-to-Estimation framework~\citep{foster2021statistical}, e.g.,
linear representability~\citep{zhu2022contextual}.

 We make the following contributions:
 \begin{itemize}
     \item We explain the problem setting (Section~\ref{sec:setting}) with careful definitions which facilitate the application of theory and reveal the unique status of expectile loss.
    \item We state the resulting algorithms (Section~\ref{sec:algorithms}), which arise via application of the Estimation-to-Decisions 
    framework~\citep{foster2021statistical}.
    \item We discuss the superior utility of expectile loss for algorithm design over more commonly used risk-measures \VaR and \CVaR (Section~\ref{sec:setting} and ~\ref{sec:algorithms}).
    \item We provide experimental support for the technique via diverse scenarios (Section~\ref{sec:experiments}). Empirically, tail control is proportionally inexpensive relative to average-case degradation, justifying the criticism of average-case guarantees in the self-tuning software literature.
 \end{itemize}


%% file: neurips2023/problemsetting.tex
\section{Problem Setting}\label{sec:setting}

This section contains tedious exposition,
necessary because (i) this work draws heavily 
on results from mathematical finance that 
cannot be presumed known by the general machine
learning audience; and (ii) careful definitions 
are key to our contribution. For the impatient
reader wanting to skip directly to
\cref{sec:algorithms}, we provide the 
following summary: \emph{use expectile loss.}  
The rest of this section answers the 
question "why?".

\paragraph{Contextual Bandits} We describe the contextual bandit
problem, which proceeds over $T$ rounds. At each round $t \in [T]$, the
learner receives a context $x_t \in \cX$ (the context space), selects
an action $a_t \in \cA$ (the action space), and then observes a loss
$l_t(a_t)$, where $l_t:\cA \to [0,1]$ is the underlying loss function.
We assume that for each round $t$, conditioned on $x_t$, $l_t$ is
sampled from a distribution $\mathbb{P}_{l_t}(\cdot\mid{}x_t)$. We
allow both the contexts $x_{1},\ldots,x_T$ and the distributions
$\mathbb{P}_{l_1},\ldots,\mathbb{P}_{l_T}$ to be selected in an arbitrary,
potentially adaptive fashion based on the history.  

\paragraph{Risk Measures} In seminal work \citet{artzner1999coherent}
presented an axiomatic approach to measuring risk.
 A risk measure is a function which maps a random
variable to $\Rinf$ and obeys certain axioms such as normalization,
translation contravariance, and monotonicity. Risk measures embed
previous approaches to measuring risk: we refer the
interested readers to \citet{meyfredi2004history}.

\paragraph{Conditional Risk-Aversion} When considering extensions of
risk-averse bandit algorithms to the contextual setting, two possible
choices are apparent: \emph{marginal} risk-aversion, corresponding
to applying a risk measure to the distribution of losses realized over the
joint context-action distribution; and \emph{conditional} risk-aversion,
corresponding to computing a risk measure on a per-context basis and then
summing over encountered contexts.  For now our focus is conditional 
risk-aversion, but after introducing terminology, we revisit the 
relationship between these two at the end of this section.

\paragraph{Contextual Bandit Regret} Conditional risk-aversion motivates
our definition of regret for finite action sets,
\begin{equation}
    \RegCB(T) \doteq \sum_{t=1}^T \mathbb{E}_{a_t}\left[ \left. \rho\left((l_t)_{a_t}\right) - \min_a \rho\left((l_t)_a\right) \right| x_t \right],
\label{eqn:condrisk}
\end{equation}
where $\rho$ is a risk measure, and the expectation is with respect to
(the algorithm's) action distribution; note $\rho$ is a function of the
adversary's loss random variable and not the realization.  For infinite
action sets we use a smoothed regret criterion: instead of competing
with the best action, we compete with any action distribution $Q$
with limited concentration $\frac{dQ}{d\mu} \leq h^{-1}$ relative to a
reference measure $\mu$,
\begin{equation}
    {\RegCB}^{(h,\mu)}(T) \doteq \sum_{t=1}^T \left(\mathbb{E}_{a_t}\left[ \left. \rho\left((l_t)_{a_t}\right) \right| x_t \right] \vphantom{\min_{ Q|\frac{dQ}{d\mu} \leq h^{-1} }} - \min_{ Q|\frac{dQ}{d\mu} \leq h^{-1} } \mathbb{E}_{a \sim Q}\left[\left. \rho\left((l_t)_a\right) \right| x_t \right]\right).
\end{equation}
Note the finite action regret is a special case, corresponding to the
uniform reference measure $\mu$ and $h^{-1} = |\cA|$.  In practice $\mu$
is a hyperparameter while $h$ can be tuned using contextual bandit
meta-learning: see experiments for details.

\paragraph{Reduction to Regression} We attack the contextual bandit
problem via reduction to regression, working with a  user-specified
class of regression functions $\mathcal{F} \subseteq (\cX \times
\cA \rightarrow [0,1])$ that aims to estimate a risk measure
$\rho$ of the conditional loss distribution.  We make the following
realizability assumption\footnote{\citet{foster2020adapting} demonstrate
misspecification is tolerable, but we do not complicate the exposition
here.}, $$
\begin{aligned}
\forall a \in \cA, t \in [T]: \exists f^* \in \mathcal{F} : f^*(x_t, a) = \rho\left(\left(l_t\right)_a\right),
\end{aligned}
$$ i.e., our function class includes a function which correctly estimates
the value of the risk measure arising from any action $a$ in context
$x_t$.  This constrains the adversary's choices, 
as  $l_t$ must be consistent with realizability,
but there are many random variables that achieve 
a particular risk value.

\paragraph{Motivation for $\EVaR$} We describe additional desirable
properties of a risk measure which ultimately determine our
choice of risk measure.  A \emph{law-invariant} risk measure is
invariant to transformations of the random variable that preserve
the distribution of outcomes, i.e., is a function of distribution
only~\citep{kusuoka2001law}. An \emph{elicitable} risk measure
can be defined as the minimum of the expectation of a loss function.
Because our algorithm operates via reduction to regression, we require
an elicitable risk measure.  A \emph{coherent} risk measure satisfies
the additional axiom of convexity: coherence is desirable because it
implies risk reduction from diversification.  To avoid confusion, note
the convexity of a risk measure is with respect to stochastic mixtures
of random variables, i.e., $\forall t \in [0, 1]: \rho(t X + (1 - t)
Y) \leq t \rho(X) + (1 - t) \rho(Y)$. For elicitable risk measures,
this is a distinct property from the convexity of the elicitation loss.

\citet{ziegel2016coherence} shows the class of elicitable law-invariant
coherent risk measures for real-valued random variables is precisely Entropic Value at Risk 
($\EVaR_q$) for $q \in \left(0, \frac{1}{2}\right]$, defined as
\begin{equation}
    \EVaR_q(D) =  \arg \min_{\hat{v} \in [0, 1]} \mathbb{E}_{v \sim D}\left[ (1 - q) \left(\left(v - \hat{v}\right)_+\right)^2\\
    +  q \left(\left(\hat{v} - v\right)_+\right)^2 \right],
\label{eqn:evar}
\end{equation}
where $(x)_+ = \max\left(x, 0\right)$. This asymmetrical strongly convex loss encourages overprediction relative to the
mean, implying infrequent large losses correspond to increased risk. A
minimizer of equation~\eqref{eqn:evar} is called an \emph{expectile}.
Certain technical qualifications are necessary for the minimum to be
achieved (bounded realization suffices).  We refer to the elicitation
loss function as \emph{expectile loss}.

\EVaR is less familiar to the machine learning community than
\VaR or \CVaR, but is a popular risk-measure in financial
applications~\citep{bellini2017risk}, whose proponents
champion the superior finite-sample guarantees induced by strong
convexity~\citep{rossello2022performance}.
\citet{waltrup2015expectile}
reveal connections between \EVaR and the risk measures \VaR and \CVaR;
in particular noting that both \VaR and \CVaR can be computed from
\EVaR.\footnote{The relationship involves differences which induces
ambiguous curvature and is therefore not viable for incorporating \VaR
or \CVaR into decision-to-estimation.} See Section~\ref{sec:algorithms}
for additional commentary.

When $q \in \left(\frac{1}{2}, 1\right)$, $\EVaR_q$ is risk-seeking.
While not our focus, the analysis remains valid therefore we state results
in terms of $\min(q, 1 - q)$.

\paragraph{Regression Oracle} We assume access to an online regression oracle \AlgReg, which is an algorithm for sequential predication under strongly convex losses using $\mathcal{F}$ as a benchmark class. More specifically, the oracle operates in the following protocol: at each round $t \in [T]$, the algorithm receives a context $x_t \in \cX$, makes a prediction $\hat{f}_t$, where $\hat{f}_t(x_t, a)$ is interpreted as the prediction for action $a$, and then observes an action $a_t \in \cA$ and realized outcome $l_t(a_t) \in [0, 1]$ and incurs instantaneous expectile loss $$
    g_t(\hat{f}_t) \doteq \left( \vphantom{q \left(\left(\hat{v} - v\right)_+\right)^2} \right. (1 - q) \left(\left(v - \hat{v}\right)_+\right)^2\\
    + q \left. \left. \left(\left(\hat{v} - v\right)_+\right)^2 \right) \right|_{v=l_t(a_t),\hat{v}=\hat{f}_t(x_t, a_t)}.
$$

We assume \AlgReg guarantees that for any (potentially adaptively chosen) sequence ${(x_t, a_t, l_t)}_{t=1}^T$,
\begin{equation}
    \sum_{t=1}^T \left( g_t(\hat{f}_t) - g_t(f^*) \right) \leq \RegEVaRq(T), \label{eqn:regevarqrealization}
\end{equation}
for some (non-data-dependent) function $\RegEVaRq(T)$.  Online regression is well-studied with many known algorithms in various cases, e.g., for linear $\mathcal{F}$ on the $d$-dimensional hypersphere, online Newton step achieves $\RegEVaRq(T) = O\left(\frac{d}{\min(q, 1 - q)} \log(T)\right)$~\citep{hazan2007logarithmic}. Furthermore, for any finite $\mathcal{F}$ we can achieve $\RegEVaRq(T) = O(\frac{1}{\min(q, 1 - q)} \log \left|\mathcal{F}\right|)$ using Vovk's aggregation algorithm~\citep{vovk1998game}.  Section 2.3 of \cite{foster2020beyond} has a more complete list of oracles.

\paragraph{Optimization Oracle} We assume an approximate (possibly randomized) optimization oracle $\AlgOpt: \mathcal{F} \times \Delta(\cA) \times \R^+ \to \Delta(\cA)$ which guarantees
$$
    \forall \hat{f} \in \mathcal{F}: \mathbb{E}_{\hat{a} \sim \AlgOpt(\hat{f}, \mu, \delta)} \left[ \vphantom{\left[\max \left(0,\hat{f}(\hat{a}) - \hat{f}(a)\right)\right]} 
    \mathbb{E}_{a \sim \mu}\left[\max \left(0,\hat{f}(\hat{a}) - \hat{f}(a)\right)\right] \right] \leq \delta,
$$ i.e., given an (estimated reward) function $\hat{f}$ the optimization
oracle can find an approximate minimizer $\hat{a}$ w.r.t the reference
measure $\mu$.  For finite action sets we can compute $\AlgOpt$
in $O(|A|)$ for all $\mu$ with $\delta = 0$.  For infinite action
sets we can compute $\AlgOpt$ with high probability via the empirical argmin over
$O\left(\frac{1}{\delta}\right)$ i.i.d. samples from $\mu$, independent
of the cardinality or dimensionality of the action space.  Of course
specific function classes may admit superior customized strategies.

\paragraph{Marginal vs. conditional, revisited} Now consider an
oblivious stationary environment where $(x, l)$ is drawn from a
fixed joint distribution $D$: further assume a law-invariant risk measure
to ease exposition, i.e., assume $\rho$ is a function of distribution only.
Marginal risk-aversion regret for a policy $\pi: X \to
\mathbb{P}(\mathcal{A})$ over a policy class $\Pi$ is defined as $$
\begin{aligned}
\text{Reg}_{\text{Marg}}(\pi) &\doteq \rho\left(D_{\text{Marg}}(\pi)\right) - \min_{\pi \in \Pi} \rho\left(D_{\text{Marg}}(\pi)\right)
\end{aligned}
$$ where $D_{\text{Marg}}(\pi)$ is defined via $$
\begin{aligned}
\mathbb{E}_{z \sim D_{\text{Marg}}(\pi)}\left[f(z)\right] &\doteq \mathbb{E}_{\substack{(x, l) \sim D \\ a \sim \pi(x)}}\left[f\left(l_a\right)\right].
\end{aligned}
$$ Marginal risk-aversion does not correspond to the expectation of
a per-context function, because the risk measure is a function of the
complete distribution.  Thus, if we apply an online-to-batch conversion
to a conditional risk-aversion regret guarantee, we end up with a
regret guarantee with respect to the expected conditional risk under
$D$ rather than the marginal risk.  For coherent risk measures,
minimizing expected conditional risk minimizes an upper bound on
marginal risk, which is sensible.  However this is unlike the
risk-neutral setting, where an adversarial guarantee provides a
tight stochastic guarantee.  In financial parlance, an algorithm designed
for the stochastic case could benefit from diversification opportunities
across context.  However, conditional risk-aversion is the appropriate
metric for scenarios where re-distributing risk across contexts is not
acceptable, e.g., software quality-of-service guarantees where the
contexts are customers.

\paragraph{Conditional risk alternative} For conditional risk there
is a plausible alternative definition.  Equation~\eqref{eqn:condrisk}
is defined by averaging the per-action risk over the policy action
distribution, but another quantity of interest is the risk measure of
the complete conditional (on context) action distribution.
Due to coherence of the risk measure, the definition in
equation~\eqref{eqn:condrisk} upper bounds this alternative, $$
\mathbb{E}_{a_t}\left[\rho((l_t)_{a_t})|x_t\right] \geq \rho(D(l_t,
a_t|x_t)),
$$
where $D(l_t, a_t|x_t)$ is the joint distribution of the action and loss
under the algorithm's conditional action distribution.  Fortunately,
unlike the marginal vs. conditional case, this is tight because 
we are competing with the best single action and the bound is tight
for degenerate distributions.  Thus optimizing our regret also controls
the risk measure of the complete conditional action distribution.

%% file: neurips2023/algorithms.tex
\begin{figure*}[t]
\begin{center}
\begin{minipage}[t]{0.46\textwidth}
\centering
\begin{algorithm}[H]
    \caption{Finite Action Set}
    \label{alg:discrete}
    \begin{algorithmic}[1]
            \FOR{$t = 1, 2, \dots, T$}
            \STATE Receive context $x_t$.
            \STATE $\hat{f}_t \leftarrow \AlgReg.\predict(x_t)$.
            \STATE $\hat{a}_t \leftarrow \AlgOpt(\hat{f}_t, 0)$.
            \STATE Sample $a_t \sim \abelong(\hat{f}_t, \hat{a}_t)$.  \label{line:expectileabelong}
            \STATE Play $a_t$ and observe loss $l_t$.
            \STATE Call $\AlgReg.\update(x_t, a_t, l_t)$.
            \ENDFOR
    \end{algorithmic}
\end{algorithm}
\end{minipage}
\hfill
\begin{minipage}[t]{0.46\textwidth}
\centering
\begin{algorithm}[H]
    \caption{Infinite Action Set}
    \label{alg:infinite}
    \begin{algorithmic}[1]
            \FOR{$t = 1, 2, \dots, T$}
            \STATE Receive context $x_t$.
            \STATE $\hat{f}_t \leftarrow \AlgReg.\predict(x_t)$.
            \STATE $\hat{a}_t \leftarrow \AlgOpt\left(\hat{f}_t, \frac{1}{4 \theta \gamma}\right)$.
            \STATE Sample $a_t \sim \contabelong(\hat{f}_t, \hat{a}_t)$.  \label{line:expectilecontabelong}
            \STATE Play $a_t$ and observe loss $l_t$.
            \STATE Call $\AlgReg.\update(x_t, a_t, l_t)$.
            \ENDFOR
    \end{algorithmic}
\end{algorithm}
\end{minipage}
\end{center}
\caption{(Left) Finite action set with exact optimization oracle.  (Right) Infinite action set with approximate optimization oracle.  Hyperparameters $\mu$ and $h$ are elided to facilitate comparison.}
\end{figure*}

\section{Algorithms}\label{sec:algorithms}

Proofs are elided to the supplemental. The proof technique has useful generality, e.g. enables the use of an approximate
minimizer in the continuous case.

By using the Estimation-to-Decision framework, we derive the resulting algorithm, which is the \emph{}{first risk-averse contextual bandit with an online guarantee}. We present two versions for finite and infinite action sets.

\subsection{Finite Action Set}

Algorithm~\ref{alg:discrete} states the finite action version of our
algorithm.  It is the \squarecb algorithm~\citep{foster2020beyond}
instantiated with an expectile loss regression oracle.
\begin{restatable}{theorem}{squarecbthm} \label{thm:discrete}
Algorithm~\ref{alg:discrete} guarantees $\RegCB(T) \leq O\left(\frac{1}{\theta} \sqrt{|\cA| T \RegEVaRq(T)}\right)$, where $\theta = \min(q, 1 - q)$.
\end{restatable}\paragraph{Proof} See Appendix~\ref{app:squarecbbound}.

We emphasize this regret is with respect to the risk measure of the best action for each context, and includes the exploration activity of the algorithm.  The $\theta$ factor indicates the difficulty of competing with an extreme expectile.  The result is intuitive as $\theta$ is the strong convexity parameter of the expectile loss.  The distribution in line~5 of Algorithm~\ref{alg:discrete} is $$
\abelong(\hat{f}_t, \hat{a}_t) = \begin{cases}
\frac{1}{|\cA| + 4 \theta \gamma \left(\hat{f}(a) - \hat{f}(\hat{a}_t)\right)} & a \neq \hat{a}_t \\
1 - \sum_{a \neq \hat{a}_t} \frac{1}{|\cA| + 4 \theta \gamma \left(\hat{f}(a) - \hat{f}(\hat{a}_t)\right)} & a = \hat{a}_t
\end{cases}.
$$

\begin{remark}
\VaR and \CVaR are alternative popular risk measures that differ from
\EVaR: \VaR lacks coherence, and \CVaR is not elicitable (only jointly
elicitable)~\citep{fissler2016higher}. Both \VaR and \CVaR do not have
strongly convex elicitation losses and hence are not compatible with
the decision-to-estimation framework.
\end{remark}

\begin{remark}
It is possible to obtain a regret bound which 
depends upon the loss of the optimal predictor
$(L^*)$ by eliciting expectile loss 
via asymmetric KL divergence combined with
a \fastcb-style 
reduction~\citep{foster2021efficient}. It is difficult to envision a
realistic risk-averse scenario in which $L^*$ is expected to be small,
i.e., in which the risk measure is expected to obtain small values yet
average case guarantees are insufficient, so we have neglected this
direction in this paper.  However in a risk-seeking scenario small $L^*$
is plausible and of potential interest.
\end{remark}

\subsection{Infinite Action Set}

Algorithm~\ref{alg:infinite} states the infinite action version of our
algorithm.  It is the \smoothcb algorithm~\citep{zhu2022contextual}, adjusted to allow for approximate minimization and instantiated with expectile loss.

\begin{restatable}{theorem}{smoothcbthm} \label{thm:infinite}
Algorithm~\ref{alg:infinite} guarantees $\RegCB^{(h, \mu)}(T) \leq O\left(\frac{1}{\theta} \sqrt{\frac{1}{h} T \RegEVaRq(T)}\right)$, where $\theta = \min(q, 1 - q)$. 
\end{restatable}\paragraph{Proof} See Appendix~\ref{app:smoothcbbound}.

The distribution in line~5 of Algorithm~\ref{alg:infinite} is $$
\begin{aligned}
\contabelong(\hat{f}_t, \hat{a}_t) &= \left(1 - \tilde{M}(\cA)\right) 1_{a=\hat{a}} + \tilde{M}, \\
\frac{d\tilde{M}}{d\mu}(a) &= \frac{1}{1 + 4 \theta \gamma h \max\left( 0, \hat{f}(a) - \hat{f}(\hat{a}) \right)}.
\end{aligned}
$$

\begin{remark}
The strong convexity of expectile loss admits other infinite action
strategies for specialized function classes, e.g., linearly structured
action spaces~\citep{zhu2022contextual}.  Relative to squared loss, expectile loss
introduces no computational overhead, and the statistical overhead is
$\min(q, 1 - q)$ due to the reduction in the strong convexity parameter.
\end{remark}


%% file: neurips2023/experiments.tex
\section{Experiments}\label{sec:experiments}

Our experiments emphasize scenarios where average-case guarantees are inadequate, and are intended to exhibit a trade-off between maximizing average-case and minimizing worst-case outcomes. 

Table~\ref{tab:datasets} in Appendix~\ref{sec:appendix_dataset} gives an overview of the scenarios and associated datasets. None of the datasets used contain either personally identifying information or offensive content. The selected datasets present various risks, such as overestimating prices in dynamic pricing, incurring unnecessary inventory costs in inventory management, and selecting worse-than-baseline configurations in self-tuning software. These risks lead to undesirable outcomes such as no-sale, financial losses, or software performance issues. To prevent these adverse outcomes, one may make a trade-off between average learning performance and the likelihood of these worst-case outcomes.

Across many domains, we found comparing a risk-averse setting with $q = 0.2$ and the risk-neutral technique with $q = 0.5$ exhibited a clear tradeoff. Note that $q = 0.5$ is the same as using the standard squared loss function. We emphasize that the choice of $q$ in practice is exogenous to the algorithm and is determined by the decision maker's level of risk-aversion.

When describing experiments, we will
use a reward convention when it is more natural, despite the analysis
using loss convention.  We will also describe experiments using the
natural reward range rather than explicitly transforming to $[0, 1]$.
In our first experiment we assess realized online expectiles directly,
but in subsequent experiments we focus on key metrics whose control is
a consequence of risk-aversion.

Continuous action experiments are implemented in PyTorch, using
Lebesque reference measure $\mu$; selecting $h$ adaptively via
\corral~\citep{agarwal2017corralling}; and computing $\hat{a}$ via
the empirical minimum over $\gamma$ samples from $\mu$.  Finite action
experiments are implemented in Vowpal Wabbit~\citep{langford2007vowpal}.
Hyperparameters are tuned using best of 59 random trials.  Confidence
intervals are 95\% coverage bootstrap intervals of online performance.
Code to reproduce all results, along with the ``Query Opt'' dataset, is available at \repourl.  
All experiments run comfortably on a commodity laptop.

\subsection{Dynamic Pricing}\label{exp:dynamicpricing}

\paragraph{Prudential} Our first dataset is from the Prudential Life
Insurance Asssessment Competition, which contains customer features along
with an associated discrete integral risk level between 1 and 8 inclusive.
We convert this to a dynamic pricing simulation as follows.  First, the
algorithm is asked to predict a risk level given the customer features.
It is assumed that the risk level is associated with a price quote which,
when correctly assessed, leads to maximum profit.  If the algorithm
overpredicts the risk level, the reward is 0; this corresponds to quoting
the customer too large of a premium and losing business to a competitor
(``no sale'').  If the risk level is not overpredicted the reward is a
linear function of the difference between the predicted and actual risk
level; this corresponds to charging too little for the premium.
Denoting the ground truth label as $y$
and the predicted label as $\hat{y}$, we have $
\mathrm{Profit}(y, \hat{y}; \beta) = \left(1 - \beta \left(y - \hat{y}\right)\right) 1_{y \geq \hat{y}}.
$ We use $\beta = 0.1$ in our experiments.

\begin{table}[b]
  \centering
  \caption{Dynamic pricing results.}
  \label{tab:dynamicpricing}
  \begin{tabular}{lcccc}
    \toprule
    Dataset & Learn $q$ & $\EVaR_{0.2}$ (\$) & Profit (\$) & No Sale (\%) \\
    \midrule
    \multirow{2}{*}{King}       & 0.2 & [18.2, 18.7] & [26.3, 26.7] &  [8.8, 9.1] \\
                                & 0.5 & [17.2, 17.6] & [28.0, 28.4] &  [17.5, 18.1] \\
    \cmidrule(r){1-1}
    \multirow{2}{*}{Perth}      & 0.2 & [22.2, 22.5] & [29.6, 29.9] & [9.5, 9.9]   \\
                                & 0.5 & [18.0, 18.5] & [31.0, 31.4] & [23.3, 23.8] \\
    \cmidrule(r){1-1}
    \multirow{2}{*}{Prudential} & 0.2 & [41.4, 41.7] & [53.4, 53.8] & [0.05, 0.09] \\
                                & 0.5 & [38.7, 39.4] & [60.6, 61.2] & [16.4, 17.0] \\
    \bottomrule
  \end{tabular}
\end{table}

\paragraph{Housing Datasets} Our next two datasets are King County and
Perth home prices, both of which contain home features along with a ground
truth listing price.  We convert these to a dynamic pricing simulation
as follows.  The algorithm must choose a listing price, and if it is
lower than the ground truth listing price, the algorithm receives the
chosen listing price as reward; if the algorithm chooses higher than the
ground truth the house does not sell and the algorithm receives 0 reward
(``no sale'').  Denoting the ground truth listing price $y$ and the
chosen listing price $\hat{y}$, we have $\mathrm{Profit}(y, \hat{y}) =
\hat{y} 1_{y \geq \hat{y}}$.  We treat (normalized) prices as continuous
actions on $[0, 1]$ and utilize Algorithm~\ref{alg:infinite} with Lebesque
reference measure.  For our regressor class, we first predict $\hat{z}:
X \to [0, 1] \times (0, \infty)$ using a linearized Cauchy kernel
machine~\citep{rahimi2007random}, and then induce a prediction function
$\hat{f}$,
\begin{equation}
\hat{f}(x, a) = a \frac{\erf\left(\frac{1 - \hat{z}_0(x)}{\hat{z}_1(x)}\right) - \erf\left(\frac{a - \hat{z}_0(x)}{\hat{z}_1(x)}\right)}{\erf\left(\frac{1 - \hat{z}_0(x)}{\hat{z}_1(x)}\right) + \erf\left(\frac{\hat{z}_0(x)}{\hat{z}_1(x)}\right)}.
\label{eqn:housing}
\end{equation}
This functional form is inspired by a truncated Gaussian random
variable, but does not imply any particular generative model.
It is simply a suitable function which is easy to implement in Pytorch.

\paragraph{Online Performance} Figure~\ref{fig:prudential} shows multiple
realized marginal expectiles on the Prudential dataset when the algorithm
is either risk-averse or risk-neutral.  This figure deviates from our
theoretical analysis in two ways.  First, it displays realized marginal
expectiles (i.e., expectiles computed from the actual sequence of rewards
experienced online) rather than summed conditional expectiles.  Second,
it extrapolates results to expectiles not optimized by the algorithm.
Nonetheless, the result exhibits the desired tail control and the
extrapolation is reasonable.

Complete results are in Table~\ref{tab:dynamicpricing}.  All CIs in
the table are computed from the online realizations.  In particular,
$\EVaR_{0.2}$ is the empirical marginal expectile experienced by the
algorithm.  We see that learning with risk-aversion ($q = 0.2$) trades
average performance (profit) for tail control.  Furthermore,
learning with risk-aversion reduces the frequency of no sale in exchange
for a reduction in profit.  Fractionally, reduction in profit is less
than the reduction in the frequency of no sale.

\paragraph{Approximate vs. exact $\hat{a}$} Unimodality of
equation~\eqref{eqn:housing} allows us to compare an approximate
maximizer, computed over $\gamma$ samples from $\mu$; with an exact
maximizer, computed using Brent's method.  Table~\ref{tab:approxvsexact}
compares on the Perth dataset. Statistically results are
similar.  Computationally, Brent's method is slower as it is not
vectorized.\footnote{Training with Brent's method is circa 2x slower on
an author's laptop, but this is problem dependent.}
\begin{table}[H]
\centering
\caption{Approximate vs. exact minimization}
\vspace{4pt}
\label{tab:approxvsexact}
\begin{tabular}{lccccc}
\toprule
Dataset & Learn $q$ & Exact? & $\EVaR_{0.2}$ & Profit (\$) & No Sale (\%) \\
\midrule
\multirow{4}{*}{Perth} & \multirow{2}{*}{0.2} & Y & [21.7, 22.1] & [29.5, 29.8] & [8.5, 8.8] \\
                       &                      & N & [22.2, 22.5] & [29.6, 29.9] & [9.5, 9.9] \\
                       & \multirow{2}{*}{0.5} & Y & [18.0, 18.6] & [31.3, 31.7] & [23.4, 24.0] \\
                       &                      & N & [18.0, 18.5] & [31.0, 31.4] & [23.3, 23.8] \\
\bottomrule
\end{tabular}
\end{table}

\begin{figure}
\centering
\begin{subfigure}[t]{.48\textwidth}
  \vskip 0pt
  \centering
  \begin{tabular}{lccc}
    \toprule
    Dataset & $q$ & Profit (\$) & Sold Out (\%) \\
    \midrule
    \multirow{2}{*}{Chicago} & 0.2 & [2.1,2.3] & [50.0,50.7] \\
                             & 0.5 & [3.7,3.9] & [20.0,20.4] \\
    \cmidrule(r){1-1}
    \multirow{2}{*}{DC} & 0.2 & [7.8,8.1] & [30.7,31.6]  \\
                        & 0.5 & [9.8,10.3] & [18.9,19.6] \\
    \cmidrule(r){1-1}
    \multirow{2}{*}{London} & 0.2 & [3.5,3.7] & [56.3,57.8] \\
                            & 0.5 & [4.7,5.0] & [28.1,29.1] \\
    \bottomrule
  \end{tabular}
  \caption{Online performance.}
  \label{tab:inventorymanagement}
\end{subfigure}
\hfill
\begin{subfigure}[t]{.48\textwidth}
  \vskip 0pt
  \centering
  \vspace{-4pt}
  \includegraphics[width=.96\linewidth]{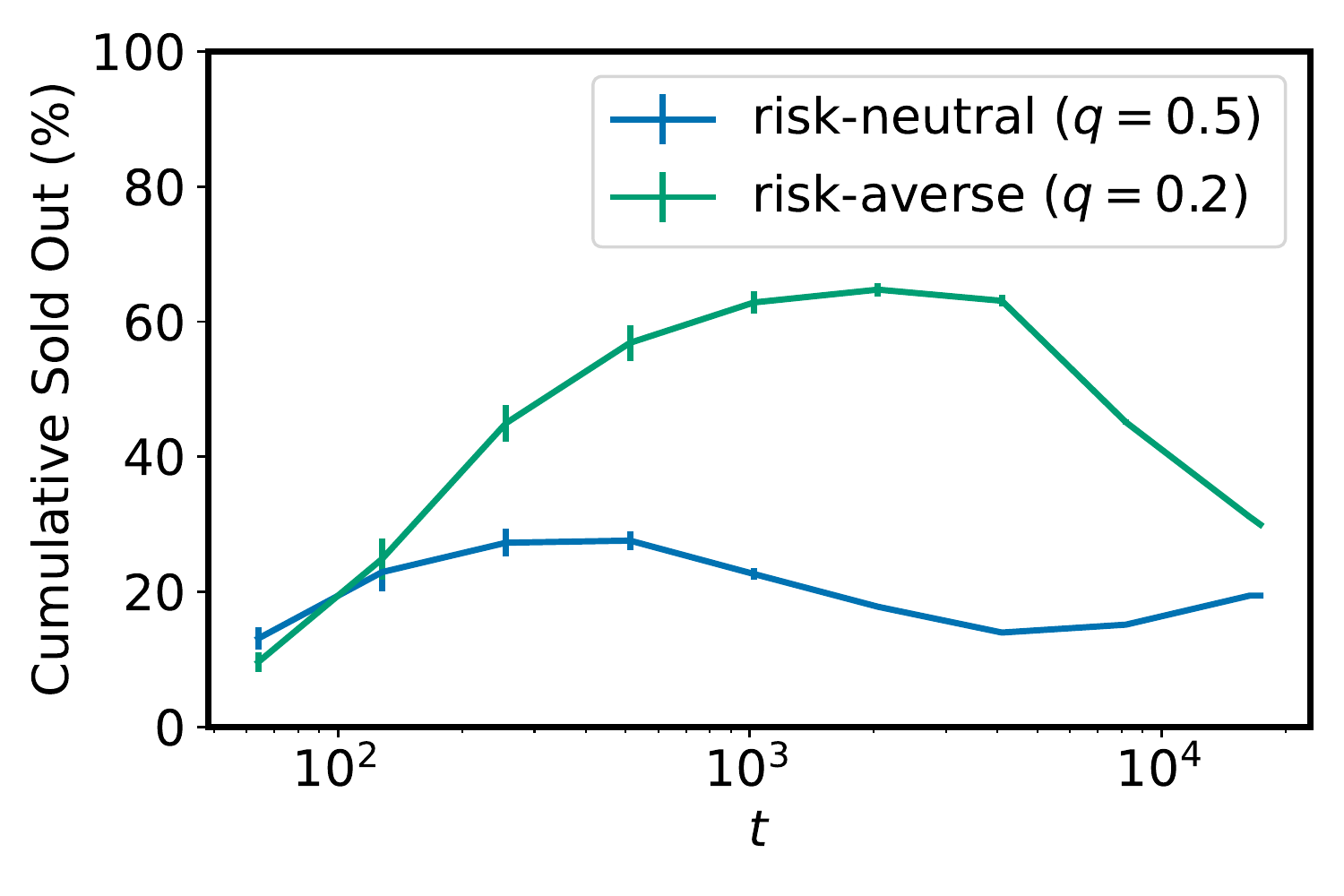}
  \vspace{-12pt}
  \caption{Cumulative sold out percentage, DC dataset.}
  \label{fig:dcsellout}
\end{subfigure}
\caption{Inventory management results.  (Left) Risk-aversion results
in lower profits but higher chance of inventory fully selling out.
(Right) Risk-aversion conservatively explores into larger allocations
from a region of safety.}
\label{fig:inventoryandsafeexplore}
\end{figure}

\subsection{Inventory Management}

\paragraph{Chicago, DC, London} Our next three datasets are public
bicycle demand datasets which contain weather and date information along
with a count of the number of bicycles demanded.  We convert these to
an inventory management simulation in which an inventory manager wants to avoid paying for inventory which is not purchased by customers. First, the algorithm
is asked to choose an allocation level given the weather and date
information.  A fixed cost per allocated bicycle is assumed.  Then, the
empirical demand level produces a fixed revenue per demanded bicycle.
We treat (normalized) bicycle allocations as continuous actions on $[0,
1]$ and allow for fractional allocation.  Denoting the ground truth
demand as $y$ and the allocation as $\hat{y}$, we have $
\mathrm{Profit}(y, \hat{y}) = \min(y, \hat{y}) - \beta \hat{y}.$

For our regressor class, we first predict $\hat{z}: X \to [0, 1] \times (0, \infty)$ using a linearized Cauchy kernel machine~\citep{rahimi2007random},
and then induce a prediction function
$\hat{f}$, $$
\hat{f}(x, a) = -\beta a + \frac{\int_0^1 \min(a, p)\ dN(p; \hat{z}_0(x), \hat{z}_1(x))}{\int_0^1 dN(p; \hat{z}_0(x), \hat{z}_1(x))},
$$
which has a (lengthy) closed form when $N(\cdot; \hat{z}_0(x),
\hat{z}_1(x))$ is a Gaussian with mean $\hat{z}_0(x)$ and variance
$\hat{z}_1(x)$.  Although inspired by a truncated Gaussian
random variable, this does not imply any particular generative model.
We use $\beta = 1/3$.

\paragraph{Online Performance} Complete results are in
Table~\ref{tab:inventorymanagement}.  All CIs in the table are computed from
the online realizations.  Learning with risk-aversion trades average
performance (profit) in exchange for a higher percentage that all allocated
inventory is demanded (sold out).  Figure~\ref{fig:dcsellout} shows
the cumulative sold out percentage as the DC dataset is consumed.
Compared to risk-neutral learning, risk-averse learning underestimates
demand and then starts to approach more accurate estimates from below.

\subsection{Self-Tuning Software}

\paragraph{Query Optimization} Our final dataset is from the exascale
cloud data processing system \scope~\citep{power2021cosmos}.
The \scope query optimizer is highly configurable and uses a
contextual bandit framework to select optimizer flags on a per-query basis~\citep{qo-advisor}. For this application, there is no single optimal flag configuration working for all input queries, and the best configuration depends on the specific query. While it is valuable to increase the overall average performance of queries, it is important to avoid regressions (queries with worse performance than the default strategy), which lead to user frustration and extra investigation work. In our experience, a non-contextual risk-neural policy only results in marginal performance lift, with a lot of regressions.

We assembled query information
(as context) and assessed the performance of multiple configurations
(as actions) per query relative to a default strategy, using fractional
change as the reward.  The number of actions per example varies depending
upon constraints imposed by the optimizer: it ranges from 2 to 22, with a
mean of 4.3 and a median of 3.  We use this dataset to construct a query
optimization simulator as follows.  First, the algorithm is presented
the query information and the configuration choices.  Then the algorithm
selects a configuration and receives the reward for that configuration.

\begin{figure}
\centering
\begin{minipage}[t]{.48\textwidth}
  \vskip 0pt
  \centering
  \includegraphics[width=.96\linewidth]{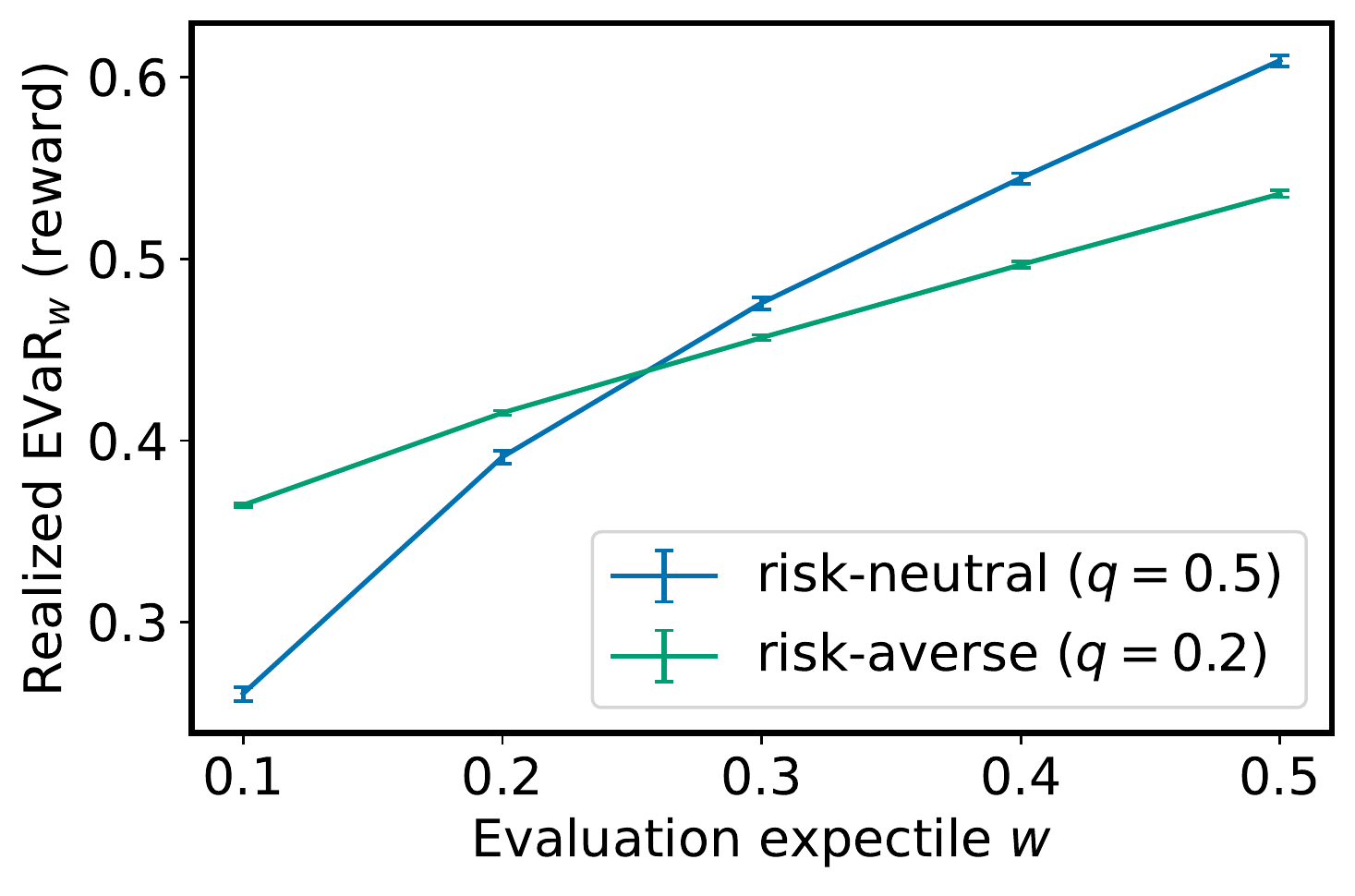}
  \caption{Realized aggregate expectiles on the Prudential dataset when the algorithm is risk-neutral $(q = 0.5)$ vs risk-averse $(q = 0.2)$.  A tradeoff between average-case guarantee and tail control is clearly evident.}
  \label{fig:prudential}
\end{minipage}
\hfill
\begin{minipage}[t]{.48\textwidth}
  \vskip 0pt
  \centering
  \vskip -1.5pt
  \includegraphics[width=.96\linewidth]{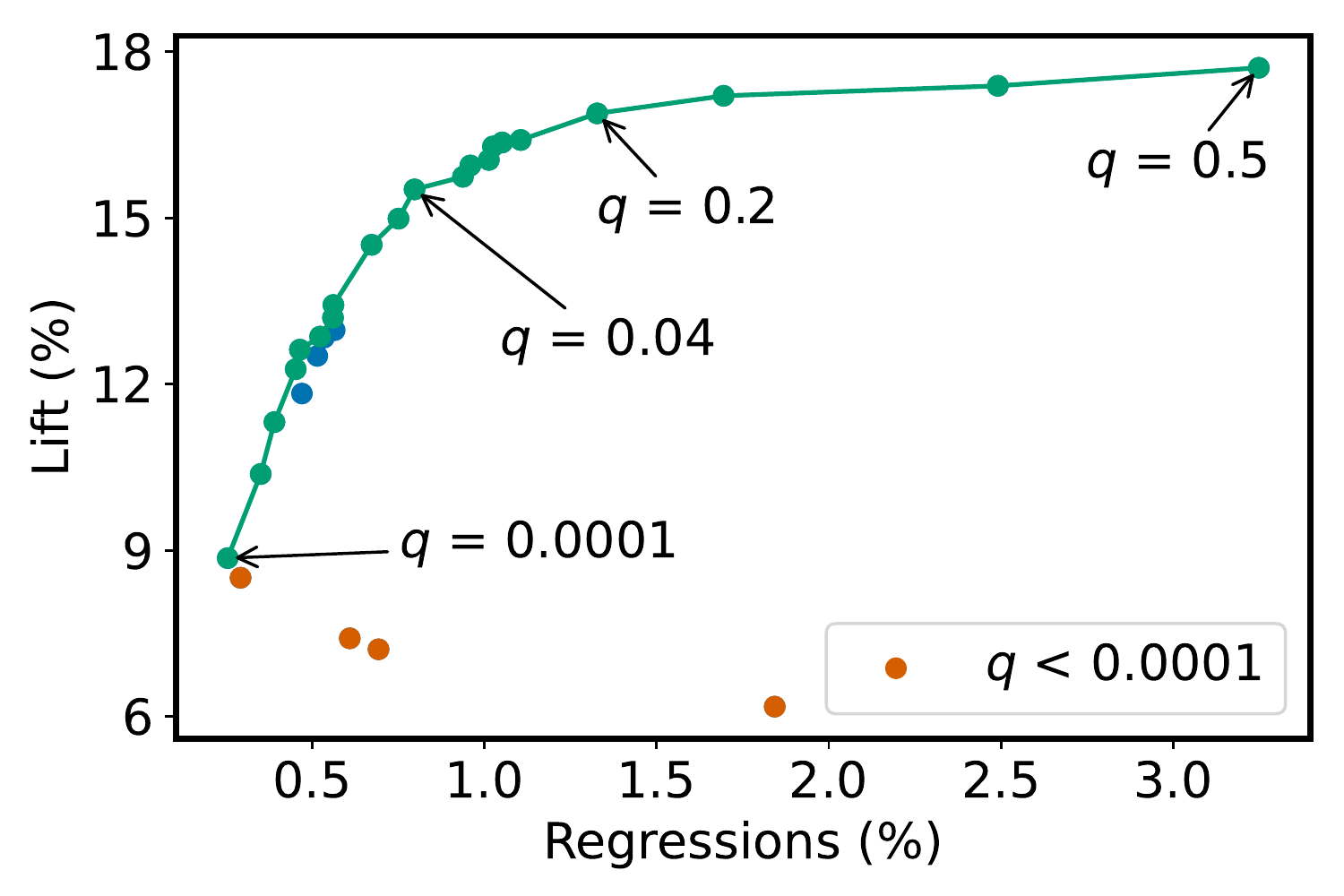}
  \caption{Query Optimization results.  Varying the learning expectile
$(q)$ yields different realized lift and regression. The Pareto front
is in green.  With moderate $q$, reductions in regression
are proportionally larger than reductions in lift.
}
  \label{fig:queryfront}
\end{minipage}
\end{figure}

Figure~\ref{fig:queryfront} summarizes the results, where the x-axis is the average performance regression for the regressed queries, and the y-axis shows the overall average performance lift.  Varying the learning
expectile $(q)$ illuminates the trade-off between lift and regression.
As seen in other experiments, there is a moderate $q$ regime where
reductions in regression are proportionally larger than reductions in lift. Comparing risk-averse $q=0.2$ with risk-neural $q=0.5$, the regressions drop by over 50\% relatively while almost maintaining the same level of lift.
For $q < 0.0001$ every point is Pareto-dominated, as anticipated by the
theoretical analysis (the regret bound degrades at extreme quantiles).

%% file: neurips2023/relatedwork.tex
\section{Related Work}\label{sec:relatedwork}


Risk-aversion has received extensive attention in the (non-contextual)
bandit literature, utilizing various risk measures.  \citet{Even-Dar2006,
Sani2012RiskAversionIM, Yu2013SampleCO, Vakili2016, zhu20d} minimize
the mean-variance, while \citet{Szorenyi2015QualitativeMB, david16,
Howard2019SequentialEO, Nikolakakis2021QuantileMB} use quantiles for
optimization.  Numerous prior works utilize Conditional Value at Risk
(\CVaR)~\citep{Tamkin2019DistributionallyAwareEF, Cardoso2019, Bhat2019,
chang2020risk, baudry21a, Khajonchotpanya2021ARA}. General risk criteria
are studied in \citet{Cassel18a, Torossian2019mathcalXArmedBO}.
\citet{Axelrod2016, aryania2021robust} consider expectiles. 
\citet{galichet2015} states algorithms for both \CVaR and the essential 
infimum. 




Prior work on risk-averse \emph{contextual} bandits is comparatively
limited. \citet{sun17a} address the adversarial contextual setting by
treating total risk as a constraint, but requires an additional risk
value observed along with cost. \citet{Bouneffouf2016ContextualBA}
presents a contextual UCB algorithm which optimizes for mean reward, but
which modulates the level of $\epsilon$-greedy exploration based upon a
risk estimate. Concurrent to our work, \citet{saux2023risk} recently also use the UCB framework solving a convex problem under the assumption of linear bandits, which do not apply to any of our non-linear predictors (e.g. Equation~\ref{eqn:housing}) in the experiments. \citet{Huang2021} study the finite sample behaviour of
off-policy estimation for a broad class of risk measures. 

The inadequacy of average-case guarantees is a recurring theme
in real-time systems applications. 
\citet{Jalaparti2013SpeedingUD} improve tail latencies of request-response workflows by minimizing variance. \citet{Schad2010RuntimeMI} use the same performance measure in cloud computing.
CVaR optimization is also present in systems applications: \citet{Mena2014} propose a multi-objective optimization technique with CVaR as risk metric in a sizing and allocation problem of renewable generation, whereas \citet{moreno2015integrating} limit risk exposure to high impact low probability events in distribution substations through this metric. However, only a small number of related bandit studies tackle risk-aware optimization in systems applications. \citet{marcus2021bao} present a bandit optimizer to improve the tail latency of queries. \citet{Sachidananda2021LearnedAF} design an autoscaler using a multi-armed bandit algorithm to optimize median or tail latency for microservice applications.

%% file: neurips2023/discussion.tex
\section{Conclusions and Future Work}\label{sec:discussion}
This paper studies the application of contextual bandits to scenarios where average-case statistical guarantees are inadequate.
We show that the composition of reduction to online regression and expectile loss is analytically tractable, 
computationally convenient, and empirically effective.  
Our experiments demonstrate the trade-off between maximizing average-case outcomes 
and minimizing worst-case performance.
These results highlight the usefulness of our method, which can
be easily applied to problems that require risk aversion.

Our reduction method exhibits an adversarial conditional risk guarantee but empirically it is also effective at controlling realized marginal risk.
However, it is possible an algorithm designed for the
stochastic case could explicitly guarantee marginal risk, e.g. via
reduction to offline reduction qua \citet{simchi2021bypassing}.

For many applications, risk-aversion is a desired end goal.  However,
explicit constraints on key metrics are also of practical interest.
Although risk-aversion implicitly controlled key metrics computed
from the complete reward distribution in our experiments, it is
complementary to approaches for constrained contextual bandits such
as ~\citet{badanidiyuru2014resourceful}.  In particular constrained
contextual bandits can control key metrics unrelated to the
reward distribution, e.g., guaranteeing quality of service while being
rewarded on cost of delivery.  Combining risk-aversion with constraints
is also a promising topic for future work.


%% file: neurips2023/appendix.tex
\appendix

\section{Regret Bound Proofs}

Note the \smoothcb bound is more general, and the finite action case a specialization.

\subsection{Proof of \smoothcb Bound} \label{app:smoothcbbound}

\smoothcbthm*
\paragraph{Proof} We have $$
\begin{aligned}
{\RegCB}^{(h,\mu)}(T) &\stackrel{(a)}{\leq} T \frac{3}{4 \min(q, 1 -q) \gamma h} + \gamma \overline{\RegEVaRq}(T), \\
&\stackrel{(b)}{\leq} T \frac{3}{4 \min(q, 1 -q) \gamma h} + 3 \gamma + 3 \gamma \RegEVaRq(T) + \frac{4 \gamma}{\min(q, 1 - q)}.
\end{aligned}
$$ where $(a)$ follows from Corollary~\ref{cor:abelong}; and $(b)$ follows from Lemma~\ref{lem:expectedregret}.  Optimizing over $\gamma$ yields $$
\begin{aligned}
{\RegCB}^{(h,\mu)}(T) &\leq \frac{1}{\min(q, 1 - q)} \sqrt{\frac{1}{h} 3 T \left(4 + 3 \min(q, 1 - q) \left(1 + \RegEVaRq(T)\right)\right)}, \\
\gamma^* &= \sqrt{\frac{3 T}{h \left(16 + 12 \min(q, 1 - q) \left(1 + \RegEVaRq(T)\right)\right)}}.
\end{aligned}
$$

\subsection{Proof of \squarecb Bound} \label{app:squarecbbound}

\squarecbthm*
\paragraph{Proof} Analogous to Theorem~\ref{thm:infinite} but using Corollary~\ref{cor:discreteabelong}, i.e., $h^{-1} = |\cA|$ and $\mu$ is the uniform distribution.

\section{Proof of convex conjugate lemma}

The following Lemma concerns bounding $$
\min_P \max_Q \max_{f^*} \mathbb{E}_{a \sim P}\left[f^*(a)\right] - \mathbb{E}_{a \sim Q}\left[f^*(a)\right] - \gamma \mathbb{E}_{\substack{a \sim P \\ l_t \sim \mathbb{P}_t}}\left[ g_t(\hat{f}_t) - g_t(f^*) \right], 
$$
i.e., the difference between contextual bandit regret and online regression regret (aka ``game value bound''), from which overall regret statements follow.

We decompose player's action distribution $P$ into two components: a sub-probability distribution $M$ which controls the adversary, and a probability distribution $N$ which distributes residual mass exploitatively.  For statistical efficiency $N$ is not needed, but for computational efficiency $N$ is useful.

\begin{lemma} Let $\phi(z)$ be a shift-invariant non-negative lower bound on the expected regret $$
\mathbb{E}_{l_t \sim \mathbb{P}_t}\left[ g_t(\hat{f}_t) - g_t(f^*) \right] \geq \phi\left(\hat{f}_t(x_t, a_t) - f^*(x_t, a_t)\right),
$$ which holds for any $\mathbb{P}_t$ s.t. $f^*(x_t, a_t)$ is a minimizer of the expected loss, let $\phi^*$ be the convex conjugate of $\phi$.  Let $\tilde{P}$ be any distribution of the form $$
\begin{aligned}
\tilde{P} &= (1 - \tilde{M}(\cA)) \tilde{N} + \tilde{M},
\end{aligned}
$$ where $\tilde{M}$ and $\tilde{N}$ are measures on the action space; $\tilde{N}(\cA) = 1$; $\tilde{M}(\cA) \leq 1$; $\tilde{M} \ll \mu$; and
\begin{equation}
\forall a: \max_{z \in \left[0, \frac{1}{h}\right]} \left(\xi\left(\frac{d\tilde{M}}{d\mu}(a), z\right) - z \left(\hat{f}(a) + \beta\right) - \kappa(a) \right) \leq 0,
\label{eqn:indifference}
\end{equation}
where
$$
\xi\left(m, z\right) \doteq \left(1 - m\right) \gamma \phi^*\left(-\frac{1}{\gamma}\right) + m \gamma \phi^*\left(\frac{1}{\gamma} \left(\frac{z}{m} - 1 \right) \right).
$$
Then $\tilde{P}$ guarantees game value bound $\left(\mathbb{E}_{a \sim \tilde{P}}\left[\hat{f}(a)\right] + \beta + \mathbb{E}_{a \sim \mu}\left[\kappa(a)\right] \right)$ when the adversary can play any distribution $Q \ll \mu$ such that $\forall a: \frac{dQ}{d\mu}(a) \leq \frac{1}{h}$.
\end{lemma}
\paragraph{Proof} Consider $P$ of the form $P = \left(1 - M(\cA)\right) N + M$ and elide $x$ dependence.
$$
\begin{aligned}
&\min_P \max_Q \max_{f^*} \mathbb{E}_{a \sim P}\left[f^*(a)\right] - \mathbb{E}_{a \sim Q}\left[f^*(a)\right] - \gamma \mathbb{E}_{\substack{a \sim P \\ l_t \sim \mathbb{P}_t}}\left[ g_t(\hat{f}_t) - g_t(f^*) \right] \\
&\leq \min_P \max_Q \max_{f^*} \mathbb{E}_{a \sim P}\left[f^*(a)\right] - \mathbb{E}_{a \sim Q}\left[f^*(a)\right] - \gamma \mathbb{E}_{a \sim P}\left[\phi\left(\hat{f}(a) - f^*(a)\right)\right] \\
&\stackrel{(a)}{=} \min_P \max_Q \mathbb{E}_{a \sim P}\left[\hat{f}(a)\right] - \mathbb{E}_{a \sim Q}\left[\hat{f}(a)\right] \\
&\qquad + \max_{z} \biggl( \mathbb{E}_{a \sim Q}\left[z(a)\right] - \mathbb{E}_{a \sim M}\left[z(a) + \gamma \phi\left(z(a)\right)\right] \\
&\qquad \qquad \qquad - \left(1 - M(\cA)\right) \mathbb{E}_{a \sim N}\left[z(a) + \gamma \phi\left(z(a)\right) \right] \biggr) \\
&\stackrel{(b)}{\leq} \min_P \max_Q \mathbb{E}_{a \sim P}\left[\hat{f}(a)\right] - \mathbb{E}_{a \sim Q}\left[\hat{f}(a)\right]
+ \mathbb{E}_{a \sim M}\left[\gamma \phi^*\left(\frac{1}{\gamma} \left( \frac{dQ}{dM}(a) - 1 \right) \right) \right] \\
&\qquad + \left(1 - M(\cA)\right) \gamma \phi^*\left(-\frac{1}{\gamma}\right) \\
&= \min_P \max_Q \mathbb{E}_{a \sim P}\left[\hat{f}(a)\right] + \beta + \mathbb{E}_{a \sim \mu}\left[\kappa(a)\right] \\
&\qquad + \mathbb{E}_{a \sim \mu}\left[-\kappa(a) - \frac{dQ}{d\mu}(a) \left(\hat{f}(a) + \beta\right) + \xi\left(\frac{dM}{d\mu}(a), \frac{dQ}{d\mu}(a)\right)\right] \\
&\leq \mathbb{E}_{a \sim \tilde{P}}\left[\hat{f}(a)\right] + \beta + \mathbb{E}_{a \sim \mu}\left[\kappa(a)\right],
\end{aligned}
$$ where $(a)$ substitutes $z(a) \doteq \hat{f}(a) - f^*(a)$; and $(b)$ is because $\left(x + \gamma \phi(x)\right)$ and $\gamma \phi^*\left(\frac{1}{\gamma} (x^* - 1)\right)$ are convex conjugates.

\begin{corollary}(Continuous Approximate Abe-Long) For $\EVaR_q$, the distribution satisfying $$
\begin{aligned}
\tilde{P} &= \left(1 - \tilde{M}(\cA)\right) 1_{\hat{a}} + \tilde{M}, \\
\frac{d\tilde{M}}{d\mu}(a) &= \frac{1}{1 + 4 \min(q, 1 - q) \gamma h \max\left( 0, \hat{f}(a) - \hat{f}(\hat{a}) \right)} \leq 1, \\
\end{aligned}
$$ where $\hat{a}$ satisfies $$
\begin{aligned}
\mathbb{E}_{a \sim \mu}\left[\max\left(0, \hat{f}(\hat{a}) - \hat{f}(a)\right)\right] &\leq \frac{1}{4 \min(q, 1 - q) \gamma},
\end{aligned}
$$ guarantees game value bound $$
\begin{aligned}
\frac{3}{4 \min(q, 1 - q) \gamma h}.
\end{aligned}
$$ 
\label{cor:abelong}
\end{corollary}
\paragraph{Proof} For $\EVaR_q$, $\phi(x) = \min(q, 1 - q) x^2$ lower bounds the expected regret by strong convexity; the convex conjugate is $\phi^*(x) = \frac{x^2}{4 \min(q, 1 - q)}$, for which $\tilde{M}(a)$ satisfies equation~\eqref{eqn:indifference} with $$
\begin{aligned}
\beta &= \frac{1 - 2 h}{4 \min(q, 1 - q) \gamma h} - \hat{f}(\hat{a}), \\
\kappa(a) &= \frac{\max\left(0, \hat{f}(\hat{a}) - \hat{f}(a)\right)}{h} + \frac{1}{4 \min(q, 1 - q) \gamma}.
\end{aligned}
$$
A bound is therefore $$
\begin{aligned}
&\frac{1}{h} \mathbb{E}_{a \sim \mu}\left[ \max\left(0, \hat{f}(a) - \hat{f}(\hat{a})\right) \right] + \frac{1}{4 \min(q, 1 - q) \gamma h} + \mathbb{E}_{a \sim \tilde{P}}\left[ \hat{f}(a) - \hat{f}(\hat{a})\right] \\
&\leq \frac{1}{2 \min(q, 1 - q) \gamma h} + \mathbb{E}_{a \sim \tilde{P}}\left[ \hat{f}(a) - \hat{f}(\hat{a})\right] \\
&\leq \frac{1}{2 \min(q, 1 - q) \gamma h} + \mathbb{E}_{a \sim \tilde{M}}\left[ \hat{f}(a) - \hat{f}(\hat{a})\right] \\
&\leq \frac{3}{4 \min(q, 1 - q) \gamma h}.
\end{aligned}
$$

\begin{corollary}(Discrete Abe-Long) For $\EVaR_q$, given a finite action set $\cA$, the distribution satisfying $$
\begin{aligned}
\tilde{P} &= \left(1 - \tilde{M}(\cA)\right) 1_{\hat{a}} + \tilde{M} \\
\tilde{M}(a) &= \frac{1}{|\cA| + 4 \min(q, 1 - q) \gamma \left( \hat{f}(a) - \hat{f}(\hat{a}) \right)} \leq \frac{1}{|\cA|},
\end{aligned}
$$ where $\hat{a}$ is an exact minimizer of $\hat{f}$, guarantees game value bound $\frac{3 |\cA|}{4 \min(q, 1 - q) \gamma}$ when competing with the best action.
\label{cor:discreteabelong}
\end{corollary}
\paragraph{Proof} Follows from above with $h^{-1} = |\cA|$, $\nu = \hat{f}(\hat{a})$, and $\mu$ uniform over $\cA$.

\section{Proof of expected regret lemma}

Our goal is to relate the total realized regret defined as $$
\begin{aligned}
\sum_{t=1}^T \left( g_t(\hat{f}_t) - g_t(f^*) \right) \doteq \sum_{t=1}^T Z_t \leq \RegEVaRq(T).
\end{aligned}
$$
to the total expected regret
$$
\overline{\RegEVaRq}(T) \doteq \sum_{t=1}^T \mathbb{E}_t\left[Z_t\right],
$$
where $\mathbb{E}_t\left[\cdot\right]$ denotes expectation conditioned $\left(\{ (x_s, a_s, l_s) \}_{s < t}, x_t, \mathbb{P}_{l_t}\right)$, i.e., averaged over the conditional action and loss distribution.

\begin{lemma}  The total expected regret is bounded by
$$
\overline{\RegEVaRq}(T) \leq 3 + 3 \RegEVaRq(T) + \frac{4}{\min(q, 1 - q)}.
$$
\label{lem:expectedregret}
\end{lemma}
\paragraph{Proof}
Note $$
M_t \doteq \sum_{s=1}^t \left(\mathbb{E}_s\left[Z_s\right] - Z_s\right) \doteq \sum_{s=1}^t \Delta M_t,
$$ is a martingale. Freedman's inequality says $$
\begin{aligned}
\mathrm{Pr}\left( M_T \geq \epsilon \right) \leq \exp\left( -\frac{\epsilon^2}{\sigma^2 + \frac{\epsilon}{3}} \right)
\end{aligned}
$$ where a.s. $\sigma^2 \geq \sum_{t=1}^T \mathbb{E}_t\left[\left(\Delta M_t\right)^2\right]$.  Integrating the tail bound, $$
\begin{aligned}
\exp\left( -\frac{\epsilon^2}{\sigma^2 + \frac{\epsilon}{3}} \right) &\leq \exp\left( -\frac{\epsilon^2}{2 \max\left(\sigma^2, \frac{\epsilon}{3}\right)} \right) \leq \exp\left(-\frac{\epsilon^2}{2 \sigma^2}\right) + \exp\left(-\frac{3}{2} \epsilon\right), \\
\mathbb{E}\left[ \left. \left| M_T \right| \{ x_t \}_{t=1}^T\right| \right] &\leq \beta_1 + \beta_2, \\
\beta_1 &\doteq \int_0^\infty \min\left(1, \exp\left(-\frac{\epsilon^2}{2 \sigma^2}\right)\right) d\epsilon \leq 2 \sqrt{\sigma^2}, \\
\beta_2 &\doteq \int_0^\infty \min\left(1, \exp\left(-\frac{3}{2} \epsilon \right)\right) d\epsilon = \frac{2}{3} + \frac{2}{3 \exp(1)} \leq 1.
\end{aligned}
$$ Thus $$
\begin{aligned}
\sum_{t=1}^T \mathbb{E}_t\left[Z_t\right] &\leq \mathbb{E}\left[\left. \sum_{t=1}^T Z_t \right| \{ x_t \}_{t=1}^T \right] + 2 \sqrt{\sigma^2} + 1 \\
&\leq \RegEVaRq(T) + 2 \sqrt{\sigma^2} + 1,
\end{aligned}
$$ where the second inequality is because the regret guarantee applies pointwise.  It suffices to bound $\sigma$.  From below we have $$
\begin{aligned}
\mathbb{E}_s\left[\left. Z_t^2 \right| a_t\right] &\leq \frac{1}{\min(q, 1 - q)} \mathbb{E}_s\left[ Z_t \right].
\end{aligned}
$$ So $$
\begin{aligned}
\sum_{t=1}^T \mathbb{E}_t\left[Z_t\right] &\leq \RegEVaRq(T) +  2 \sqrt{\frac{1}{\min(q, 1 - q)}} \sqrt{\sum_{t=1}^T \mathbb{E}_t\left[Z_t\right]}  + 1 \\
\implies \sum_{t=1}^T \mathbb{E}_t\left[Z_t\right] &\leq 3 + 3 \RegEVaRq(T) + \frac{4}{\min(q, 1 - q)}.
\end{aligned}
$$

\subsection{Bound for $\sigma$}

$$
\begin{aligned}
\mathbb{E}_t\left[(\Delta M_t)^2\right] &\leq \mathbb{E}_t\left[Z_t^2\right]  = \mathbb{E}_t\left[ \mathbb{E}_t\left[\left. Z_t^2 \right| a_t\right] \right], \\
\mathbb{E}_t\left[\left. Z_t^2 \right| a_t\right] &= \mathbb{E}_t\left[ \left. \left(g_t(\hat{f}_t) - g_t(f^*)\right)^2 \right| a_t \right].
\end{aligned}
$$ From convexity and $|\nabla g_t| \leq \max(q, 1 - q) \leq 1$, we have $$
\left|\hat{f}_t - f^*\right| \leq \left| g_t(\hat{f}_t) - g_t(f^*) \right| \leq \left|\hat{f}_t - f^*\right|,
$$ thus
$$
\begin{aligned}
\mathbb{E}_t\left[\left. Z_t^2 \right| a_t\right] &\leq \mathbb{E}_t\left[ \left. \left(\hat{f}_t - f^*\right)^2 \right| a_t \right] \\
&= \frac{1}{\min(q, 1 - q)} \mathbb{E}_t\left[ \left. \min(q, 1 - q) \left(\hat{f}_t - f^*\right)^2 \right| a_t \right] \\
&\leq \frac{1}{\min(q, 1 - q)} \mathbb{E}_t\left[ \left. Z_t \right| a_t\right].
\end{aligned}
$$

\section{Datasets}\label{sec:appendix_dataset}
\begin{table*}[h]
\begin{center}
\begin{minipage}{\textwidth}
  \caption{Datasets. Numbered names indicate OpenML ids.~\citep{vanschoren2014openml}.}
  \label{tab:datasets}
  \centering
  \begin{tabular}{llccc}
    \toprule
    Scenario & Name & License & $T$ & Actions \\
    \midrule
    \multirow{3}{*}{Dynamic Pricing} & King County $(42092)$ & CC-BY\footnote{\url{https://creativecommons.org/licenses/by/2.0/}} & 21613 & $[0, 1]$ \\
                                     & Perth $(43822)$ & CC-BY & 33656 & $[0, 1]$ \\
                                     & Prudential\citep{Prudential} & Custom\footnote{\url{https://www.kaggle.com/competitions/prudential-life-insurance-assessment/rules}} & 59381 & 8 \\
    \cmidrule(r){1-1}
    \multirow{3}{*}{Inventory Management} & Chicago\citep{Chicago} & CC-0\footnote{\url{https://creativecommons.org/share-your-work/public-domain/cc0}} & 34617 & $[0, 1]$ \\
                               & DC $(42712)$ & CC-BY & 17379 & $[0, 1]$ \\
                               & London\citep{London} & OGL\footnote{\url{https://en.wikipedia.org/wiki/Open_Government_Licence}} & 17414  & $[0, 1]$ \\
    \cmidrule(r){1-1}
    Self-Tuning Software & Query Opt & CC-BY & 48681 & Finite Variadic \\
    \bottomrule
  \end{tabular}
\end{minipage}
\end{center}
\vskip -12pt
\end{table*}